# Style Transfer Generative Adversarial Networks: Learning to Play Chess Differently

**Muthuraman Chidambaram & Yanjun Qi**
Department of Computer Science
University of Virginia
Charlottesville, VA 22903, USA
{`mc4xf,yanjun`}`@virginia.edu`

## Abstract

The idea of style transfer has largely only been explored in image-based tasks, which we attribute in part to the specific nature of loss functions used for style transfer. We propose a general formulation of style transfer as an extension of generative adversarial networks, by using a discriminator to regularize a generator with an otherwise separate loss function. We apply our approach to the task of learning to play chess in the style of a specific player, and present empirical evidence for the viability of our approach.

## 1 Introduction

Gatys et al. (2015) showed that a convolutional neural network (CNN) model could be trained to transfer the unique styles present in human art onto other images. However, the style transfer loss used in their paper, as well the losses used in the follow-up work of Ulayanov et al. (2016) and Johnson et al. (2016), were specific to image-based tasks. This makes it difficult to extend their work on style transfer to other tasks where unique human styles are present, such as playing games.

Motivated by this problem, we present a general framework for style transfer, which we term style transfer generative adversarial networks (STGANs) as an extension of the generative adversarial networks (GANs) described by Goodfellow et al. (2014). Our proposed framework consists of a generator $G$, which learns to perform a given task, and a discriminator $D$, which learns to predict whether the same task was performed in a specific style. These two models are trained in an adversarial fashion by using the discriminator to regularize the generator, so that the generator learns to perform the given task in a way that is consistent with the style designated by the discriminator.

In this paper, we examine an application of STGANs to the task of learning to play chess in the style (made precise in appendix) of a designated player. Essentially, a generator is trained to evaluate chess board positions, and is then combined with a search function to generate moves. A discriminator is trained to distinguish the moves selected using the generator from the moves of a designated player, and is used to bias the generator's evaluations towards the style of the designated player.

## 2 STGAN Model

The key difference between our proposed STGAN model and the GAN model is that the generator loss in our model is not structured purely in terms of the discriminator. Instead, the generator loss is defined to be specific to the given task, which we take to be generating optimal chess board evaluations in this paper. We then define a style transfer generator loss by regularizing the original generator loss using the discriminator.

### 2.1 Generator

We structured our generator to be similar to the Deep Pink model described by Bernhardsson (2014). The generator $G$ is thus a fully connected feedforward neural network with a 768 unit-wide input layer, two 2048 unit-wide hidden layers with ReLU activations, and a single linear output unit. The generator takes as input a chess board, which is represented as a 768 element vector corresponding to the locations of the 12 different chess pieces, and outputs a real number as an evaluation. Positive evaluations signify that the board is in white's favor, while negative evaluations signify that the board is in black's favor. We train the generator using triplets of chess boards $(x_G, y_G, r_G)$ taken from games played by top chess players, where $x_G$ is an initial board, $y_G$ is the board after a player has made a move on $x_G$, and $r_G$ is the board after a random move has been made on $x_G$. The function $G$ is learned such that $G(x_G) = G(y_G)$ and $G(y_G) > G(r_G)$ if it is white's turn to move and $G(y_G) < G(r_G)$ if it is black's turn to move. The assumption made here is that the generator is being trained on boards taken from masters' games, so board evaluations should not change much after a move has been played (neither player gives the other a significant advantage). Consequently, a random non-master move is considered to be much worse, so the board evaluation should become more positive if the random move was played by black and more negative if the random move was played by white. We formulate the generator loss $J^{(G)}(\theta_G)$ as:

$$J^{(G)}(\theta_G) = -\frac{1}{m}\sum_{i=1}^{m}[\log(\sigma(G(x_G^{(i)}) - G(y_G^{(i)}))) + \log(\sigma(G(y_G^{(i)}) - G(x_G^{(i)}))) \\ + \log(\sigma(p_i(G(y_G^{(i)}) - G(r_G^{(i)}))))] \quad (1)$$

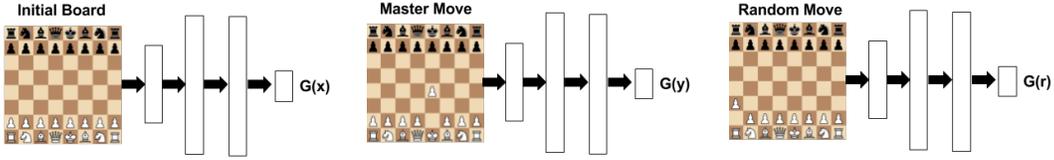

Figure 1: The generator network learns $G(x_G) = G(y_G)$, as well as $G(y_G) > G(r_G)$ if $y_G$ corresponds to a move made by white and $G(y_G) < G(r_G)$ otherwise.

Where $m$ is the batch size, $\sigma$ is the sigmoid function, and $p_i$ is 1 if it is white's turn to move on the input board, and -1 otherwise. The terms $\log(\sigma(G(x_G^{(i)}) - G(y_G^{(i)})))$ and $\log(\sigma(G(y_G^{(i)}) - G(x_G^{(i)})))$ enforce the inequalities $G(x_G) > G(y_G)$ and $G(y_G) > G(x_G)$, thereby attempting to learn $G(x_G) = G(y_G)$. The term $\log(\sigma(p_i(G(y_G^{(i)}) - G(r_G^{(i)}))))$ enforces the inequality $G(y_G) > G(r_G)$ if it is white's turn to move, and $G(y_G) < G(r_G)$ if it is black's turn to move.

## 2.2 DISCRIMINATOR

The discriminator, which learns a function $D$, is set up identically to the generator, save for a 1536 unit-wide input layer and a sigmoid output. The discriminator takes as input a valid chess move, which is represented as the concatenation of the vector representations of a pair of boards, and outputs the probability that the move was played by a designated player. Training is done using pairs of sequential boards $(x_D, y_D)$ taken from the games of a designated player, as well as fake move pairs $(x_D, M(x_D))$ generated by selecting moves using the generator $G$. The board $M(x_D)$ is chosen using the negamax search described by Campbell & Marsland (1983) with a search depth of one and the generator as the board evaluation function. The discriminator is optimized by maximizing $D((x_D, y_D))$ and minimizing $D((x_D, M(x_D)))$, which corresponds to minimizing the following discriminator loss $J^{(D)}(\theta_D)$:

$$J^{(D)}(\theta_D) = -\frac{1}{m}\sum_{i=1}^{m}D((x_D^{(i)}, y_D^{(i)})) + \frac{1}{m}\sum_{i=1}^{m}D((x_D^{(i)}, M(x_D^{(i)}))) \quad (2)$$

Here we have opted to structure the discriminator loss after the loss described by Arjovsky et al. (2017) for training Wasserstein GANs (WGANs).

## 2.3 STYLE TRANSFER

Style transfer is done by using the discriminator to regularize the generator. This is achieved by defining a style transfer generator loss $J_{ST}^{(G)}(\theta_G)$ as:

| Move | Baseline, $k=0$ | Style Transfer, $k=1$ | Style Transfer, $k=2$ |
|------|-----------------|------------------------|------------------------|
| f8e7 | 0.710 | 0.565 | 0.325 |
| d7d5 | 0.703 | 0.603 | 0.479 |

Table 1: Negamax move evaluations produced by the baseline and style transfer generator networks

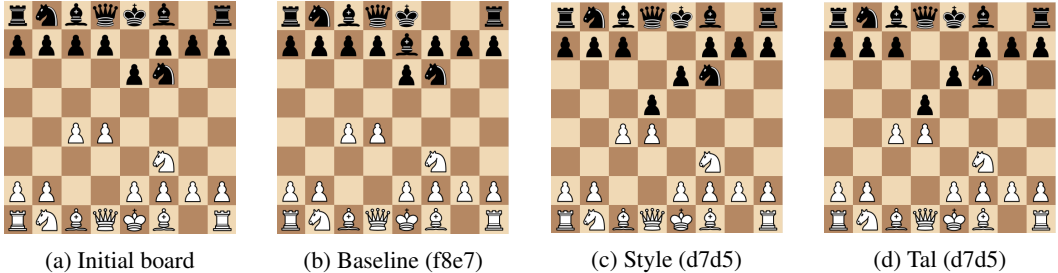

(a) Initial board  (b) Baseline (f8e7)  (c) Style (d7d5)  (d) Tal (d7d5)

Figure 2: Positions reached after queen's pawn opening sequence of moves

$$J_{ST}^{(G)}(\theta_G) = J^{(G)}(\theta_G) - \frac{1}{m} \sum_{i=1}^{m} kD((x_D^{(i)}, M(x_D^{(i)}))) \qquad (3)$$

Where $k$ is a hyperparameter that controls the level of influence the style designated by the discriminator should have on the generator. Since certain boards $x_G^{(i)}$ may not be represented in the discriminator's training data, we choose to use initial boards $x_D^{(i)}$ for the regularization term.

### 2.4 TRAINING

The discriminator and the generator are updated simultaneously by gradient descent on $J^{(D)}$ and $J_{(ST)}^{(G)}$, but the discriminator is updated 5 times for each generator update, as described by Arjovsky et al. (2017) in the WGAN paper. Examples taken from the most recent training batch of the discriminator are used for regularization in each generator update. The discriminator's weights are also clamped to be in the range $[-0.01, 0.01]$, once again consistent with the WGAN approach.

## 3 RESULTS

Training data for the generator was obtained by extracting all standard chess games played in 2016 between players with ratings above 2000 from the FICS games database. For the discriminator, we chose to predict the style of late chess grandmaster Mikhail Tal, and extracted his 2431 available games from PGN Mentor as training data.

We trained multiple generator networks with varying values of the regularization parameter $k$, with $k = 0$ being treated as the baseline. Due to the cost of having to perform a negamax search for each generated move during training, all networks were trained for 10 epochs with only 100 batches of size 64 sampled from the training data in each epoch. After training, each network was tested by generating move sequences (once again using a negamax search with depth one) in response to the queen's pawn opening sequence of moves.

Figure 2 shows the positions reached by the generator networks after white plays the queen's pawn opening sequence of moves (d2d4, c2c4, g1f3), as well as an actual position commonly reached by Tal for the same sequence. The style transfer networks end in the same position as Tal, whereas the baseline network reaches a position never played by Tal within the data. Table 1 shows the difference in move evaluations between the networks, and it can be seen that the final Tal move (d7d5) becomes more favored as $k$ is increased. It should be noted that the style transfer networks still learn that the move f8e7 is a good move in the last position (positive negamax evaluation), so they are not simply overfitting to Tal's moves.

# 4 APPENDIX

## 4.1 BACKGROUND: GENERATIVE ADVERSARIAL NETWORKS

The STGAN model is an extension of the generative adversarial network (GAN) framework put forth by Goodfellow et al. (2014). The GAN framework consists of a discriminator $D$ with parameters $\theta_D$, which attempts to determine the probability that an input is real, and a generator $G$ with parameters $\theta_G$, which attempts to generate input data from noise that is intended to "fool" the discriminator. $G$ is defined to be a differentiable function which transforms a random sample $z$ drawn from a prior distribution into an input sample $G(z)$ for the discriminator. The discriminator function $D$ is then trained to maximize the probability of input samples $x$ drawn from a data distribution $P_{data}$ and minimize the probability of generated input samples $G(z)$ by minimizing the following loss $J^{(D)}(\theta_D)$:

$$J^{(D)}(\theta_D) = -\frac{1}{m}\sum_{i=1}^{m}[\log(D(x^{(i)})) + \log(1 - D(G(z^{(i)})))] \quad (4)$$

Where $m$ corresponds to the batch size. The generator is trained simultaneously to generate samples that are indistinguishable from actual samples drawn from $P_{data}$ by minimizing the loss:

$$J^{(G)}(\theta_G) = \frac{1}{m}\sum_{i=1}^{m}\log(1 - D(G(z^{(i)}))) \quad (5)$$

## 4.2 DEFINING AND MOTIVATING STYLE TRANSFER

It is very difficult to define "styles" for games such as chess. While a human can assess a player as aggressive or defensive in a qualitative manner, we are not aware of any quantitative metrics for measuring such characteristics in chess. In this paper, we take "style" to mean a favoritism towards certain positions, specifically within opening sequences in chess. We then assess style by considering a common opening sequence of moves for the player with the white pieces and then observing the sequence of response moves played by the style transfer and baseline models. If the sequence of response moves falls into an opening repertoire commonly used by the player whose style we were attempting to transfer, we consider the style transfer to be successful. We choose to use a definition of style transfer that is based on opening sequences of moves due to the higher degree of subjectivity in measuring style within the later portions of chess games, which we think would be an interesting area of future research.

Furthermore, it is also important to understand why style transfer provides value in a game such as chess. There exist many chess-playing AIs that can play well above the grandmaster level, but we note that these AIs use a fixed play style based on tree search and position heuristics. Thus, they do not provide an easy way for players to train against different types of opponents. We believe that successfully emulating the style of specific players would provide significant pedagogical value to those looking to further their chess skills, as they would be able to practice against a variety of different openings and preferred styles. We use style transfer as opposed to simply training a neural network on the games of an individual player due to the fact that an individual player's game data does not typically cover enough board positions to train a proficient model (i.e. the model plays very poor moves in positions not represented in its data).

Finally, while our example use case of the STGAN framework is chess, we believe that the framework can be applied to other domains. For example, for image style transfer, one could construct an STGAN in which the generator is an image generation model and the discriminator is trained on images corresponding to a specific style (i.e. Picasso).

## 4.3 FURTHER EXPERIMENTS: MIKHAIL CHIGORIN

While our main results focused on transferring the style of the famous chess player Mikhail Tal to our baseline model, we also ran experiments with another notable player: Mikhail Chigorin.

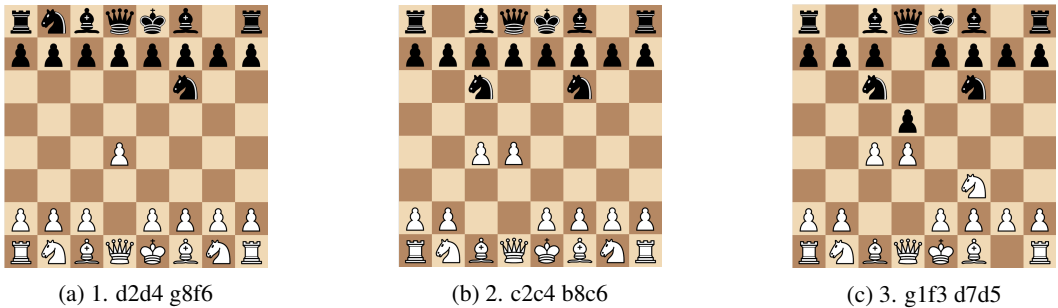

(a) 1. d2d4 g8f6     (b) 2. c2c4 b8c6     (c) 3. g1f3 d7d5

Figure 3: Response move sequence played by Chigorin style transfer model

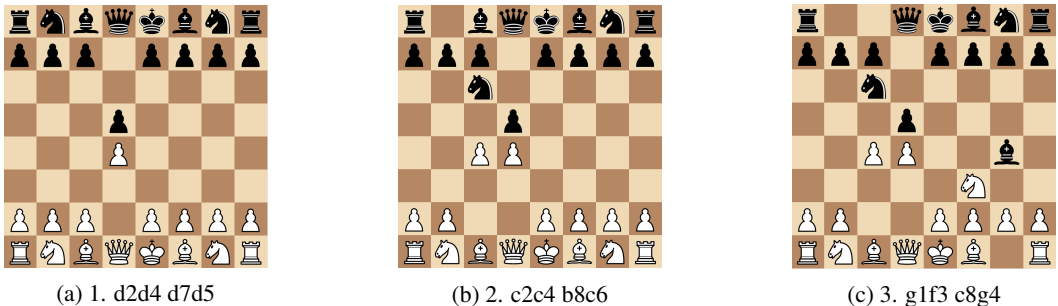

(a) 1. d2d4 d7d5     (b) 2. c2c4 b8c6     (c) 3. g1f3 c8g4

Figure 4: Classical sequence of the Chigorin defense

Chigorin is famous for popularizing a specific response to the queen's pawn opening in chess, which is the eponymous Chigorin defense (Watson, 1981). Similar to the approach we used for Tal, we extracted Chigorin's 688 available games from PGN Mentor and used them as training data for the discriminator in the STGAN framework.

Given the classical queen's pawn opening move sequence described in the main results section, the Chigorin style transfer model ($k = 1$) responded with the move sequence shown in Figure 3. The classical sequence of moves played in the Chigorin defense is shown in Figure 4. It can be seen that the final position reached by the style transfer model represents a variation of the Chigorin defense, which seems to have been a result of the baseline model heavily favoring the first move g8f6 due to its representation in the master games used for training.